\documentclass{article}
\usepackage{times}
\usepackage{graphicx}
\usepackage{amsfonts}
\usepackage{amssymb}
\newtheorem{Theorem}{Theorem}
\newtheorem{Definition}{Definition}

\newenvironment{Proof}{\par\bgroup{\it Proof.}}{~~$\bullet$\egroup\par}

\def\eq{\!=\!}

\def\p{\par}
\def\paroff{\catcode`\^^M=10}

\def\Same{\hbox{\sf same}}

\begin{document}

\title{On the Relationship between the Posterior and Optimal Similarity}
\author{Thomas M.~Breuel\\PARC, 3333 Coyote Hill Rd., Palo Alto, CA 94304, USA}
\date{November 2003\footnote{
This TR was originally written in 2003, but only submitted to Arxiv in 2007.
References have not been updated to include more recent work.
}
}
\maketitle

\paroff

\begin{abstract}

For a classification problem described by the joint density $P(\omega,x)$,
models of $P(\omega\eq\omega'|x,x')$ (the ``Bayesian similarity measure'')
have been shown to be an optimal similarity measure for nearest neighbor
classification.

This paper analyzes demonstrates several additional properties of that
conditional distribution.

The paper first shows that we can reconstruct, up to class labels, the class 
posterior distribution $P(\omega|x)$ given $P(\omega\eq\omega'|x,x')$,
gives a procedure for recovering the class labels, and gives an asymptotically
Bayes-optimal classification procedure.

It also shows, given such an optimal similarity measure, how to construct
a classifier that outperforms the nearest neighbor classifier and achieves
Bayes-optimal classification rates.

The paper then analyzes Bayesian similarity in a framework where a
classifier faces a number of related classification tasks (multitask learning)
and illustrates that reconstruction of the class posterior distribution
is not possible in general.

Finally, the paper identifies a distinct class of classification
problems using $P(\omega\eq\omega'|x,x')$ and shows that using
$P(\omega\eq\omega'|x,x')$ to solve those problems is the Bayes optimal
solution.

\end{abstract}

\section{Introduction}

Statistical models of similarity have become increasingly important in
recent work on information retrieval \cite{HofPuz98nips}, case-based
reasoning \cite{faltings97probabilistic}, pattern
recognition\cite{baxter97canonical}, and computer vision
\cite{mahamud02,MahHeb03}.

Of particular interest is Bayesian similarity, a discriminatively
trained model of $P(x \hbox{~and~} x' \hbox{~are in the same
class}|x,x')$, which we will abbreviate as $P(\Same|x,x')$.

These models have been demonstrated to work well in a number of
pattern recognition and visual object recognition problems
\cite{2001-breuel-icassp,mahamud02,MahHeb03,2003-breuel-icdar-2}.

\p

It is easy to see that nearest neighbor classification using
$1-P(\Same|x,x')$ minimizes the risk that the class labels for $x$
and $x'$ differ and therefore is optimal for 1-nearest
neighbor classification \cite{mahamud02,MahHeb03}.

However, beyond that observation, there have been several kinds of
analysis of Bayesian similarity.

The first, presented by Mahamud \cite{mahamud02,MahHeb03} is an analysis
considering a single instance of a classification problem, determined
by a joint distribution $P(\omega,x)$ of class labels $\omega$ and
feature vectors $x$.

The authors also argue for the existence of useful invariance properties
of Bayesian similarity functions when those functions have a specific
form \cite{MahHeb03}.

The second is an analysis based on a hierarchical Bayesian framework
presented by Breuel \cite{2003-breuel-icdar-2}, which effectively
considers Bayesian similarity in the context of a distribution of
related classification tasks.

\p

This paper analyzes the relationship between Bayesian similarity
$P(\Same|x,x')$ and the class posterior distribution $P(\omega|x)$ 
in both the non-hierarchical and hierarchical cases and uses those
results to construct an asymptotically Bayes-optimal classification
procedure using Bayesian similarity.

It also presents a new statistical model for the kinds of
discrimination tasks described in \cite{MahHeb03} and 
demonstrates that Bayesian similarity is the Bayes-optimal
solution for those tasks.

The implications of these results for applications of Bayesian
similarity will be discussed at the end.

\section{Bayesian Similarity}

Consider a classification problem in which feature vectors
$x\in\mathbb{X}=\mathbb{R}^n$ and class variables $\omega\in\{1,\ldots,c\}$
are jointly distributed according to some distribution $P(x,\omega)$.

\begin{Definition}

Let $P(x,\omega)$ be the distribution for a classification
problem.

Given two samples from this distribution, $(x,\omega)$ and
$(x',\omega')$, we define {\em Bayesian similarity} as the probability
$P(\omega\eq\omega'|x,x')$. 

When $\omega$ and $\omega'$ are clear from context, we will usually
denote this as $P(\Same|x,x')$.

\end{Definition}

Let $x_\omega$ be a sample that has somehow been selected as a
``prototype'' for class $\omega$.

It is natural to classify some unknown feature vector $x$ using
the rule:

\begin{equation}\label{bsim}
D(x) = \arg \max_{\omega'} P(\omega\eq\omega'|x,x_{\omega'})
\end{equation}

That is, we classify the unknown feature vector $x$ using the class
associated with the training example $x_{\omega'}$ that is most similar
to it in the sense of Bayesian similarity.

\p

Observe that Equation~\ref{bsim} is analogous to nearest neighbor
classification if we use $d(x,x') = 1-P(\omega\eq\omega'|x,x')$ as
the similarity function.

Because $P(\omega\eq\omega'|x,x')$ is, by definition, the probability 
that $x$ and $x'$ have the same class label, it is also the Bayes-optimal
misclassification rate using a nearest neighbor rule; therefore, 
nearest-neighbor classification using $d(x,x') = -P(\omega\eq\omega'|x,x')$ 
is an optimal nearest neighbor classifier.

\p

Nearest neighbor classification using $d(x,x') =
1-P(\omega\eq\omega'|x,x')$ is not necessarily Bayes-optimal; in fact,
the asymptotic bounds on its performance are no better than those known
for traditional nearest neighbor methods \cite{mahamud02,MahHeb03}.

However, when $x_\omega$ is an unambiguous prototype, that is,
$P(\omega|x_\omega) = 1$, then classification with Bayesian similarity
is Bayes optimal:

\begin{equation}
P(\omega\eq\omega'|x,x_{\omega'}) = \sum_{\omega'} P(\omega|x) P(\omega'|x')
                                = \sum_{\omega'} P(\omega|x) \delta(\omega,\omega')
                                = P(\omega'|x)
\end{equation}

That is, in the case of unambiguous training examples, Equation~\ref{bsim} just reduces to Bayes-optimal classification.

\section{Relationship between $P(\Same|x,x')$ and $P(\omega|x)$}

While we have seen some relationships between nearest neighbor
classification and Bayesian similarity in the previous section and in
the literature \cite{mahamud02,MahHeb03}, the question arises whether there
are better ways of taking advantage of $P(\Same|x,x')$ and
whether we can achieve Bayes-optimal classification using
a Bayesian similarity framework.

\p

Consider a two-class classification problem; that is, $\omega\in\{0,1\}$.

Now, examine the probability $P(\Same|x,x)$; that is, the probability
that two samples with the same feature vector actually have the same class.

This probability is not equal to 1 in general because the both of the
class conditional densities $P(x|\omega\eq 0)$ and $P(x|\omega\eq 1)$ may 
be nonzero at $x$.

We obtain:

\begin{eqnarray}\label{sqsum}
P(\Same|x,x) &=& P(\omega\eq 0|x) P(\omega\eq 0|x) + P(\omega\eq 1|x) P(\omega\eq1|x) \\
                         &=& (P(\omega\eq 0|x))^2 + (P(\omega\eq 1|x))^2 \\
                         &=& (P(\omega\eq 0|x))^2 + (1-P(\omega\eq 0|x))^2
\end{eqnarray}

We can solve this for $P(\omega\eq 0|x)$ up to a sign:

\begin{equation}\label{res}
P(\omega\eq 0|x) \,=\, \frac{1}{2} \pm \frac{1}{2} \sqrt{2 P(\Same|x,x) -1}
\end{equation}

Note that $P(\Same|x,x)\in[\frac{1}{2},1]$, so this is well-defined
and real.

Given $P(\Same|x,x')$, in particular, we have
$P(\Same|x,x)$, and from Equation~\ref{res}, we see that we can
reconstruct $P(\omega\eq 0|x)$ up to a single choice of a sign at each point.

Of course, while this gives us a lot of information about $P(\omega|x)$,
the unknown sign is crucial for classification.

\p

Now consider the decision regions for the minimum error decision rule:

$D_0 = \{ x | P(\omega\eq 0|x)>\frac{1}{2} \}$ and 
$D_1 = \{ x | P(\omega\eq 0|x)<\frac{1}{2} \}$.

That is, given an unknown feature vector $x$,
we decide $\omega\eq 0$ when $x\in D_0$ and $\omega\eq 1$ when
$x\in D_1$.

If $x$ is not contained in either decision region, we can
make an arbitrary choice between classes $0$ and $1$.

\p

\def\half{\frac{1}{2}}
\def\myeps{d}

Now consider two points $x$ and $x'$.

Assume they both come from $D_0$:

Then, for some positive $\myeps$ and $\myeps'$, $P(\omega\eq 0|x) = \half+d$ and 
$P(\omega\eq 0|x') = \half+d'$.

Therefore,

\begin{eqnarray}\label{eqsame}
P(\Same|x,x')&=&(\half+\myeps)(\half+\myeps')+(\half-\myeps)(\half-\myeps')\\
	    &=&\half+2\myeps\myeps'
\end{eqnarray}

If both come from $D_1$, the result is the same.

If one comes from $D_0$ and the other comes from $D_1$, then, for some
positive $\myeps$ and $\myeps'$,

\begin{eqnarray}
P(\Same|x,x')&=&(\half+\myeps)(\half-\myeps')+(\half-\myeps)(\half+\myeps')\\
	    &=&\half-2\myeps\myeps'
\end{eqnarray}

Since the $\myeps$ and $\myeps'$ are both positive, this means that if $x$ and
$x'$ are in the same decision region, $P(\Same|x,x')>\half$, and otherwise
$P(\Same|x,x')<\half$.

Therefore, for any two points $x$ and $x'$, we can decide whether they
are in the same decision region by seeing whether $P(\Same|x,x')>\half$.

\p

Using these two results, we can now state the following theorem:

\begin{Theorem}
We can reconstruct either $P(\omega|x)$ or $1-P(\omega|x)$ from $P(\Same|x,x')$.
\end{Theorem}

\begin{Proof}
Compute the two possible values for $P(\omega\eq 0|x)$ using Equation~\ref{res}.

Pick a point $x$ at which $P(\omega\eq 0|x)\neq\half$, i.e., where
$P(\Same|x,x')\neq\half$.

$P(\omega\eq 0|x)$ is then either less than $\half$ or greater than $\half$.

Arbitrarily pick one of these; this is a choice of membership
of $x$ in $D_0$ or $D_1$.

Use the constraint $P(\omega|x,x')>\half$ for points in the same decision
region to assign all other points to decision regions.

Given the decision regions and the values from Equation~\ref{res},
we have reconstructed either $P(\omega|x)$ or $1-P(\omega|x)$, depending
on whether our arbitrary choice above was correct or not.
\end{Proof}

This means that if we have an estimate of the Bayesian similarity 
function $P(\Same|x,x')$, we have already identified the class posterior
distribution up to a choice of two: $P(\omega|x)$ (the correct class posterior
distribution), and $1-P(\omega|x)$.

\p

Once we have $P(\Same|x,x')$, training samples only serve to distinguish
the two possibilities for the reconstructed class posterior distributions.

Since the prior probability for either choice is $\half$, we can determine
which of the two possibilities applies by considering the ratio of the 
probability of the samples given the models.  

That is, if we write $P_A(\omega|x)$ and $P_B(\omega|x)$ for the two
possibilities, then we evaluate

\begin{equation}\label{lratio}
r = \frac{\prod_i P_A(\omega_i,x_i)}{\prod_i P_B(\omega_i,x_i)}
  = \frac{\prod_i P_A(\omega_i|x_i)P(x_i)}{\prod_i P_B(\omega_i|x_i)P(x_i)}
  = \frac{\prod_i P_A(\omega_i|x_i)}{\prod_i P_B(\omega_i|x_i)}
\end{equation}

If $r>1$, then $P_A$ is the more likely possibility, otherwise 
$P_B$ is the more likely possibility.

\p

So, if we take this together, we have a Bayes-optimal classification
procedure given $P(\Same|x,x')$ and a set of prototypes or samples
$(\omega_i,x_i)$:

first, we compute the two possible values of $P(\omega|x)$ at each point
using Equation~\ref{res}, then we use Equation~\ref{eqsame} to assign
those values to the two possible branches, and then finally use the
prototypes to identify which of the two branches is the more likely
using Equation~\ref{lratio}.

Finally, we classify using the reconstructed class conditional
distribution $P(\omega|x)$.

\p

The only purpose that training samples obtained in addition to the
Bayesian similarity function $P(\Same|x,x')$ serve in this procedure is
to determine which of the two possible choices of the reconstructed
$P(\omega|x)$ is the correct one.

Asymptotically, the above procedure for making the choice between
the two possibilities, can be seen to be correct with probability one.

Therefore, this classification procedure is asymptotically Bayes-optimal.

\p

Compare that with the proposed use of Bayesian similarity in a nearest
neighbor classification procedure.

First, the approach described above is very different from a nearest
neighbor classifier, because it integrates information from all
samples.

Second, given $P(\Same|x,x')$ and labeled training examples, a nearest
neighbor classifier using Bayesian similarity, even asymptotically, is not
guaranteed to come within more than a factor of two of the Bayes-optimal
error rate \cite{mahamud02,MahHeb03}, while the procedure described above
will almost always reach the Bayes-optimal error rate.

\section{Multi-Class Case}

The previous section showed that for one large class of classification
problems (namely, two-class classification problems), knowledge
of the Bayesian similarity function is essentially equivalent to
knowledge of the class posterior distributions.

That already demonstrates that, given $P(\Same|x,x')$, 1-NN classification
is not an admissible classification procedure (i.e., there is a procedure
that is uniformly better).

However, while it is not central to the main argument, it is an 
interesting question to ask whether that approach 
generalizes to the multi-class case.

Let us sketch the argument here without making a full, formal
proof.

\p

As before,

\begin{equation}\label{condind}
P(\Same|x,x') = \sum_i P(\omega=i|x) P(\omega=i|x')
\end{equation}

Now, assume that are looking at $c$ classes and $n$ points $x_j$
and write $p_{ij} = P(\omega=i|x_j)$.  

Also, write $s_{ij}$ for $P(\Same|x_i,x_j)$.

Then, we have 

\begin{equation}\label{sqrel}
s_{ij} = \sum_k p_{ki} p_{kj}
\end{equation}

The $s_{ij}$ are $\half n (n-1)$ given quantities, and
there are $(c-1)\,n$ unknown quantities $p_{ij}$.

We have enough equations to solve for the unknowns 
when $n \geq 2c-1$.

\p

Of course, as in the two-class case, given any solution $p_{ij}$,
any permutation of class labels remains a solution, and as before,
this is expressed as an uncertainty of signs in the system of equations
given by Equation~\ref{sqrel}.

But, as in the two-class case, there is only a finite number of
possibilities, and we can distinguish among them by computing the
likelihoods of the actual set of training samples for each of the
different possible solutions.

Therefore, we see that, as in the two-class case, we can reconstruct
the class conditional density up to permutation. 

As before, any additional training examples or prototypes we use merely
serve to pick the most likely possibility among this finite set.

\section{Batched Hierarchical Bayesian Similarity}

In the previous sections, we have seen that knowledge of the
Bayesian similarity function $P(\Same|x,x')$ is mostly equivalent
to knowledge of the class posterior distribution $P(\omega|x)$.

In effect, Bayesian similarity is a suboptimal application 
of $P(\omega|x)$.

This raises the question of whether using Bayesian similarity for nearest
neighbor classification is of any use at all.

Both this and the next section answer that question in the affirmative.

While Bayesian similarity is not useful for simple classification problems,
it is useful for hierarchical Bayesian problems and actually
Bayes-optimal for certain discrimination problems.

In fact, all previous applications of Bayesian similarity in the literature,
including \cite{MahHeb03} are probably better analyzed in one of
these two frameworks than as simple classification problems.

\p

One way of understanding learning similarity measures for nearest neighbor
classifiers is to think of the problem as learning a similarity
measure for a collection of related task.

For example, in an OCR problem, a similarity function might generally
be able to evaluate the similarity of different character shapes to one
another, but when applied to a specific classification problem, the
identity of individual characters is given by a set of training examples.

See \cite{heskes98,2002-breuel-icpr1} for further information.

The idea of a collection of related classification problems can be formalized
in its most general form as that of hierarchical Bayesian methods.

After describing hierarchical Bayesian classification, we will return
to its relationship with Bayesian similarity.

\p

In a hierarchical Bayesian framework, we assume that the distribution
governing the classification problem is parameterized
by some parameter vector $\theta$, which is itself distributed according
to some prior $P(\theta)$.

We write $P_\theta(x|\omega)$ or, equivalently, $P(x|\omega,\theta)$
for the parameterized class conditional density.

If we are just given individual samples from such a hierarchical Bayesian
model, the model is merely a particular representation of a non-hierarchical
density using an integral \cite{Berger80}:

\begin{equation}\label{hbint}
    P(x|\omega) = \int P(x|\omega,\theta) P(\theta) d\theta
\end{equation}

\p

In a batched hierarchical Bayesian problem, a classifier
faces a collection of batches, where the samples $(\omega_i,x_i)$ within
each batch are drawn using the same parameter $\theta$.

The Bayes-optimal classification for a batch of samples 
$B = \{\ldots,(\omega_i,x_i),\ldots\}$
can be derived from the class conditional density for that batch:

\begin{equation}\label{eqhier}
    P(x|\omega) = \int \prod_i P(x_i|\omega_i,\theta) P(\theta) d\theta
\end{equation}

Note that this differs from a non-batched hierarchical Bayesian model,
for which the class conditional density for the same batch would be
$P(x|\omega) = \prod_i \int P(x_i|\omega_i,\theta) P(\theta) d\theta$.

\p

Let us now return to the question of how a hierarchical Bayesian approach
relates to Bayesian similarity.

Trivially, we have

\begin{equation}
    P(\Same|x,x') = \int P(\Same|x,x',\theta) P(\theta) d\theta
\end{equation}

This function can be approximated by taking pairs of samples $(\omega,x)$ and
$(\omega',x')$ from the same batch $\theta$ and training a classifier
with it.

We refer to this as {\em batched training}.

That is, it is learned analogously to Bayesian similarity in the
non-hierarchical cases, but all pairs of feature vectors $x$ and $x'$
used for training are taken from the same batch.

\p

What is the equivalent to Equations~\ref{condind} and~\ref{sqrel}?

Those equations relied on the relationship $P(\omega\eq\omega'|x,x') 
= P(\omega|x) P(\omega|x')$.

But the equivalent relationship is not true in the hierarchical Bayesian
case. 

While $P(\omega\eq\omega'|x,x',\theta) = P(\omega|x,\theta)
P(\omega|x',\theta)$, the same is not true in general for the
corresponding marginal distributions after integration over $\theta$:
$P(\omega\eq\omega'|x,x') \stackrel{\scriptsize ?}{=} P(\omega|x)
P(\omega|x')$.

Therefore, given $n$ sample points $x_1,\ldots,x_n$, in general,
we may have to estimate the values for all $c^n$ combinations of
classifications $P(\omega_1,\ldots,\omega_n|x_1,\ldots,x_n)$,
and for that, the $\half n (n-1)$ Bayesian similarity values
$P(\Same|x_i,x_j)$ do not provide sufficient information in general.

\p

Therefore, for hierarchical Bayesian classification, knowledge of the
Bayesian similarity is not, in general, equivalent to knowledge of the
class posterior distributions.

However, even in a hierarchical Bayesian framework, it is still true 
that Bayesian similarity is the optimal similarity function for nearest
neighbor classification:

for any $1-P(\omega\eq\omega'|x,x') = 1-\int P(\omega\eq\omega'|x,x',\theta)
P(\theta) d\theta$ is the risk that the class labels associated with $x$ and
$x'$ differ, and minimizing that risk minimizes the overall risk of
misclassification in a 1-nearest neighbor framework (this is the analogous
argument to that made in \cite{mahamud02,MahHeb03}).

We can therefore state:

\begin{Theorem}
    Batch-trained Bayesian similarity is the optimal distance function
    for 1-nearest neighbor classification in a batched hierarchical
    Bayesian classification problem.
\end{Theorem}

\section{Discrimination Tasks}

In the previous section, we looked at a hierarchical Bayesian classification
task.

Let us now look at a closely related problem.

\p

Mahamud \cite{mahamud02,MahHeb03} considers the problem of
determining whether two image patches in different images come
from the same object or different objects.

For this, they train a Bayesian similarity model $P(\Same|x,x')$ and
use it to make this decision for real images.

\p

They analyze this by formulating it as a non-hierarchical
classification problem and postulate an underlying joint distribution
$P(\omega,x)$ between class labels and feature vectors.

That presupposes that some class structure exists over the
image patches; that is, that image patches can be classified into
a fixed set of categories and that the purpose of nearest neighbor
classification is to recover those categories.

But the authors do not demonstrate that such a class structure
actually exist, and its existence does not appear particularly
plausible.

\p

Consider, for example, feature vectors consisting of color histograms
over image patches.

While it is meaningful to ask whether two such color histograms
are sufficiently similar between two images to have come from
the same object, there is no obvious classification of color
histograms that is independent of the specific problem instance.

\p

\def\Diff{\hbox{\sf different}}

There are two different condition, the $\Same$ condition and the
$\Diff$ condition.

Let us write $S=1$ and $S=0$ for the two conditions, respectively.

Under the $S=1$ condition, two unknown feature vectors $x$ and $x'$ 
are produced by the same patch, parameterized as $\theta$.

Under the $S=0$ condition, two unknown feature
vectors are produced by different patches, parameterized as 
$\theta$ and $\theta'$.

The task Mahamud \cite{mahamud02,MahHeb03} set out to solve is whether a given
pair of feature vectors $x$ and $x'$ was produced under the
$S=1$ or $S=0$ conditions.

In order to solve this problem, they postulate the existence
of an underlying classification problem $P(\omega,x)$ and then
address it using non-hierarchical Bayesian similarity.

Their justification for using Bayesian similarity is that
$\omega$ is unobservable, so training a traditional classifier
would be impossible.

\p

If we don't invoke an underlying, unobservable class structure,
how should we analyze this kind of discrimination problem?

Let us say that the possible surface patches on a 3D object are
parameterized by some parameter vector $\theta$.

Furthermore, let the viewing parameters for that surface patch
be given as $\phi$ and that there is some random noise variable $\nu$.

Then, the feature vector representing the appearance of the surface 
patch in the image, for unknown viewing parameters and noise, is given by 

\begin{equation}
    P(x|\theta) = \int P(x|\theta,\phi,\nu) P(\phi) P(\nu) d\phi d\nu
\end{equation}

The problem is now to determine whether two samples $x$ and $x'$ come
from the same distribution $P(x|\theta)$.

\p

For concreteness, let us write down the distributions involved in this
problem.

The class conditional density under the $\Same$ condition is

\begin{equation}
P(x,x'|S=1) = \int P(x|\theta) P(x'|\theta) P(\theta) d\theta
\end{equation}

For the $S=0$ condition, it is given by 

\begin{equation}
P(x,x'|S=0) = \int P(x|\theta) P(\theta) d\theta \int P(x'|\theta') P(\theta') d\theta' = P(x) P(x')
\end{equation}

The joint distribution is just the mixture:

\begin{equation}
    P(x,x',S) = P(x,x'|S=1) P(S=1) + P(x,x'|S=0) P(S=0)
\end{equation}

Applying Bayes rule gives us 

\begin{equation}
    P(S=1|x,x') = \frac{P(x,x'|S=1) P(S=1)}{\sum_{S\in\{0,1\}} P(x,x',S)}
\end{equation}

Nowhere in this derivation of the posterior distribution was it
necessary to postulate an underlying class structure.

Furthermore, if we obtain a model of $P(S|x,x')$ from training
data and use it for deciding whether $x$ and $x'$ were generated
under $S=0$ or $S=1$ conditions, our decision procedure
will be Bayes-optimal because $P(S|x,x')$ is the optimal discriminant
function for $S$.

Suboptimality of the use of $P(S|x,x')$ for classification was a result
of the fact that in classification, we are trying to make a decision 
about $\omega$, not $S$.

\section{Discussion}

In this paper, we have seen three distinct uses of Bayesian similarity:
as a similarity measure for non-hierarchical classification problems, 
as a similarity measure for batched hierarchical classification problems,
and as a similarity measure for discrimination tasks.

\p

The paper has shown that for non-hierarchical classification problems,
models of $P(\Same|x,x')$ are equivalent to models of $P(\omega|x)$,
up to permutation of the class labels.

That makes the use of Bayesian similarity for individual classification
problems merely a variation of learning a classifier.

In a sense, $P(\Same|x,x')$ is too problem specific: it ``knows so
much'' about the particular classification problem $P(\omega,x)$ that
we might as well use $P(\omega|x)$ directly.

Although this paper did not show it formally, that is likely to
be a problem with any optimal similarity measure for nearest neighbor
classification.

\p

Intuitively, what we would like is a similarity measure that works
well across an entire class of related problems.

We can formalize this notion of a class of related problems in a
hierarchical Bayesian framework
\cite{Berger80,baxter97canonical,heskes98,2002-breuel-icpr1}.

When we consider Bayesian similarity in such a framework, it is not
equivalent to knowledge of the class posterior distributions anymore.

However, the property that it is an optimal similarity function for
nearest neighbor classification remains.

This means that in a hierarchical Bayesian setting, Bayesian
similarity is a procedure that is distinct from other methods
and may have useful applications;

unlike more direct or generative implementations of hierarchical
Bayesian models \cite{heskes98,2002-breuel-icpr1}, Bayesian similarity
models appear to be easier to implement and train.

It is important to remember that such hierarchical models are trained
differently from the non-hierarchical models: for non-hierarchical models,
samples $x$ and $x'$ used for training $P(\Same|x,x')$ are taken from
the entire distribution, while for hierarchical models, such samples
are only taken from within a batch that was sampled using the same
distributional parameters $\theta$.

\p

Finally, the paper has presented a novel analysis of $P(\Same|x,x')$
for discrimination tasks like those considered in
\cite{mahamud02,MahHeb03} and demonstrated that the use of
$P(\Same|x,x')$ in such tasks is, in fact, Bayes-optimal.

This is an important result because those kinds of discrimination
class are quite common in computer vision applications.

\bibliographystyle{plain}
\bibliography{nips-nn}
\clearpage

\end{document}